\documentclass[letterpaper, 10 pt, journal, twoside]{IEEEtran}
\usepackage{amsmath,amsfonts}
\usepackage{algorithmic}
\usepackage{algorithm}
\usepackage{array}
\usepackage[caption=false,font=normalsize,labelfont=sf,textfont=sf]{subfig}
\usepackage{textcomp}
\usepackage{stfloats}
\usepackage{url}
\usepackage{verbatim}
\usepackage{graphicx}
\usepackage{cite}
\usepackage{lipsum}
\usepackage{bm}
\usepackage{booktabs}
\usepackage{xcolor}
\usepackage{balance}

\begin{document}

\title{Mobile Pedipulation for Object Sliding via Hierarchical Control on a Wheeled Bipedal Robot}

\author{Yue Qin, Yulun Zhuang, Zelin Shen, and Yanran Ding
\thanks{The authors are with the Department of Robotics, University of Michigan, Ann Arbor, MI 48109, USA.
        {\tt\footnotesize \{yueqin, yulunz, zelin, yanrand\}@umich.edu} }
\thanks{
This work was supported by National Science Foundation, under grant 2427036 and Schaffler Group USA, Inc.
}
}



\maketitle
\pagestyle{empty}
\thispagestyle{empty}

\begin{abstract}
In this letter, we present a hierarchical control framework that enables wheeled bipedal robots to perform planar object sliding tasks with their wheeled legs. 
The proposed approach formulates a nonlinear model predictive controller (NMPC) based on a reduced-order three rigid bodies (TRB) dynamical model that explicitly accounts for the hip roll degree of freedom and multiple wheel-environment contact modes, which is essential for lateral stepping and pedipulation tasks. 
Within this framework, the NMPC simultaneously regulates robot locomotion and interaction forces, allowing the robot to stably execute both rolling and object manipulation behaviors.
A trajectory-optimization-based robot-object motion planner is developed to generate reference motions that incorporate stick-slip transitions in ground-object contact.
Two representative pedipulation motions, namely scooting and lateral sliding, are validated through real-world hardware experiments, in which the robot successfully retrieves a $1~kg$ object from under a desk and slides a $4~kg$ object over a distance of $0.228~m$ via scooting. 
\end{abstract}

\begin{IEEEkeywords}
Legged robot, wheeled robot, optimization and optimal control, nonprehensile manipulation
\end{IEEEkeywords}

\section{Introduction}
Wheeled bipedal robots have the potential to perform versatile tasks since they integrate the mobility of the wheels and the maneuverability of the legs.
Various recent works have demonstrated the good terrain adaptability of wheeled bipedal robots for locomotion tasks, such as rolling~\cite{klemm2020lqr,yu2023modeling,yu2025whole,wang2025enhanced,zhao_design_2024,wang2024design}, stepping~\cite{xin2020online}, and jumping~\cite{chen2020underactuated,zhuang2021height,klemm2019ascento}. 
However, the versatility of wheeled bipedal robots remains far below that of their bipedal siblings.
In particular, the ability to dynamically interact with the environment and external objects has not been demonstrated on wheeled bipedal robot platforms.

Leveraging the legs for manipulation, or \textit{pedipulation}, offers unique advantages for certain specialized tasks compared with approaches that isolate manipulation from locomotion using a mounted manipulator~\cite{arm_pedipulate_2024, lin2024locoman,zhang_motion_2020}. 
Wheeled bipedal robots provide a promising platform for pedipulation because their wheel-type end effectors allow the robot to maintain mobility without periodically making and breaking ground contact for balance, offering a stable contact interface to apply forces to objects~\cite{mason_mobile_1999,srinivasa2002experiments}.
This capability makes them especially suitable for planar object sliding manipulation, where the wheels are used to slide the object.
Such sliding-based object interaction offers a practical solution for mobile manipulation scenarios involving non-prehensile and non-pushable objects, such as welcome mats, cardboard sheets lying flush with the ground, or objects trapped in corners.

In this letter, a hierarchical control framework is presented that enables a wheeled bipedal robot to pedipulate planar objects via sliding.
The key component in the framework is a nonlinear model predictive controller (NMPC), which handles object interaction and balances the robot within a unified formulation.
Within the NMPC, a three rigid bodies (TRB) reduced-order model is adopted, capturing the interaction dynamics between the torso and the wheels without imposing additional assumptions on the wheel contact or workspace. The TRB model is further extended to TRB-Object (TRBO) model by incorporating a point-mass object model and Coulomb friction complementarity constraints, enabling robot-object interaction planning in both motion planner and NMPC.

\begin{figure}[tb]
    \centering
    \includegraphics[width=0.96\linewidth]{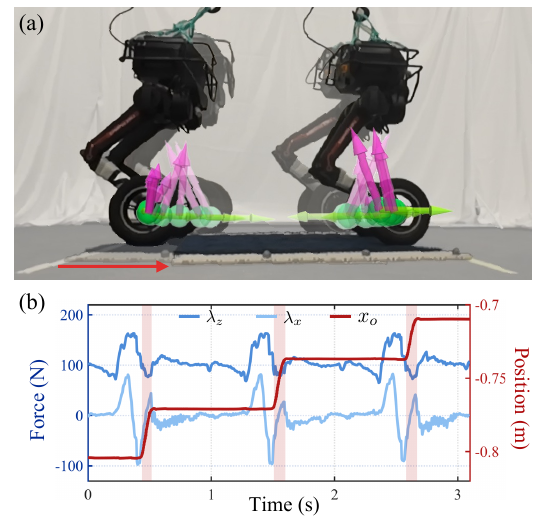}
    \caption{The scooting pedipulation experiment. (a) Snapshots of the first and last cycles of the scooting motion. The red arrow indicates the object displacement during the experiment. The green spheres and arrows represent the wheel's position and velocity in the NMPC prediction horizon, while the pink arrows indicate the predicted GRF. Entities fade with increasing prediction time. (b) Data plots of the third to the fifth scooting cycles. $\lambda_z$ and $\lambda_x$ denote the vertical and horizontal GRF acting on the robot, respectively, and $x_o$ represents the object position. The red shaded regions indicate the time intervals during which the object slides.}
    \label{fig:scooting}
\end{figure}

The proposed framework is validated on \textit{Tron1} wheeled bipedal robot platform. Locomotion skills are validated through velocity tracking and stepping experiments. This hierarchical control framework enables pedipulation skills where the robot can initiate object sliding motion in both longitudinal and lateral directions. Building upon these skills, the robot successfully retrieves an object trapped under a desk by combining stepping and lateral sliding. In simulation, the \textit{Tron1} could initiate the sliding of an object up to $23$ kg on a slippery surface with a friction coefficient of $0.3$, highlighting its capability in heavy-load tasks. The extended TRBO-based NMPC enables accurate object pedipulation even in the presence of model mismatch.

The main contributions include:

\begin{itemize}
    \item A novel TRB reduced-order model that captures the hip roll degree of freedom (DoF) and admits non-rolling contact modes without imposing a non-slip assumption.
    \item First demonstration of dynamic non-prehensile pedipulation on a wheeled bipedal robot with stick-slip-aware planning.
    \item Hardware validation, including non-trivial tasks such as under-desk retrieval that require lateral-sliding skill.
\end{itemize}

\section{Related Work}
Planar sliding manipulation using robotic manipulators has been extensively studied and is closely related to the task considered in this work.
Most existing approaches rely on the assumption that the interaction wrench on the manipulator end effector can be directly and precisely controlled, simplifying the manipulator dynamics in order to alleviate the computational burden of the underlying contact-aware optimization~\cite{yi2023precise,yang2024dynamic,howe1996practical}.
While such an assumption applies to fully actuated manipulators, it does not hold in dynamic pedipulation, where the robot is highly underactuated, and the reaction forces are not directly controllable.

This challenge has motivated recent work on loco-manipulation for legged robots, where the controller must jointly account for underactuated robot dynamics and robot-object interaction. 
For example, Sleiman et al.~\cite{sleiman2021unified} proposed a unified MPC framework for quadrupedal loco-manipulation by augmenting the robot centroidal dynamics with object dynamics. Li et al.~\cite{li2023dynamic} utilized a single rigid body model for humanoid loco-manipulation control, in which object interaction is represented as external forces. 
Although these reduced-order formulations are effective for conventional legged platforms, they are not directly applicable to wheeled bipeds because wheel-ground contact is governed by distinctive nonholonomic rolling constraints. 
This motivates a control framework that explicitly models wheeled-biped dynamics and rolling-contact constraints for pedipulation.

Model-based predictive control is a common control strategy for wheeled bipedal robots, which exhibit inherent non–minimum-phase dynamics, introducing undershoot behaviors~\cite{klemm2020lqr,yu2023modeling,yu2025whole,wang2025enhanced,zhao_design_2024,wang2024design,xin2020online,chen2020underactuated}.
Utilizing reduced-order models for the predictive controller is proven effective~\cite{ding2021representation, ding2019real, zhuang2025kinodynamic,di2018dynamic}.
Klemm et~al.~\cite{klemm2020lqr} utilized a wheeled inverted pendulum model, achieving high-performance motion control on the \textit{Ascento} robot.
Chen et~al.~\cite{chen2020underactuated} proposed the wheeled-spring-loaded inverted pendulum model for the jumping motion planning of wheeled bipedal robots. 
While these simplified models capture the sagittal-plane dynamics using a planar wheeled–pendulum–like approximation, their representable motions are only valid when the wheeled bipedal robot operates around a symmetric configuration.

To capture the 3D dynamics of wheeled bipedal robots, Yu et~al.~\cite{yu2023modeling} proposed a wheeled rigid-body dynamics model in which the torso and wheels are modeled as separate floating rigid bodies. 
The resulting dynamics are linearized around a nominal configuration to formulate a linear MPC controller.
Wang et~al.~\cite{wang2025enhanced} adopted a similar rigid-body-based modeling approach within a NMPC framework. 
Nevertheless, both models assume no-slip rolling contact, which prevents them from capturing alternative wheel–environment contact modes, such as non-contact and slipping, limiting their applicability to walking and sliding motions.
Moreover, both works simplify the hip as a single-DoF pitch joint and omit the hip roll joint, which is important for lateral stepping and object-interaction behaviors.
A general reduced-order model that captures nonlinear dynamics while explicitly accounting for the hip roll joint and multiple contact modes has not been established for wheeled bipedal robots. 
In this work, we address this gap by proposing a TRB reduced-order model for wheeled bipedal robots without imposing additional constraints on the wheel contact or workspace. An NMPC controller is formulated based on the TRB model to provide robust and versatile control for both locomotion and pedipulation tasks.

\section{Preliminaries}
\subsection{Model Predictive Control}
MPC is an optimization-based control strategy that computes the control input by repeatedly solving a finite-horizon optimal control problem (OCP). At each time step, only the first control input is applied, and then the horizon is shifted forward in a receding horizon manner. The general form of MPC for a discrete-time system is formulated as:
\begin{equation}
    \begin{aligned}
    \min_{\bm{x}, \bm{u}} \quad 
    &\ell_T(\bm{x}_N) + \sum_{k=0}^{N-1} \ell(\bm{x}_k, \bm{u}_k)\\
    \text{s.t.}\, \quad 
    & \bm{x}_{k+1} = \bm{f}(\bm{x}_k,\bm{u}_k) \\
    & \bm{g}(\bm{x}_k,\bm{u}_k) \leq0\\
    & \bm{g}_T(\bm{x}_N) \leq 0 \\
    & \bm{x}_0 = \bm{x}_{\text{op}} \\
    & k =0,\dots,N-1,
    \end{aligned}
\end{equation}
where the $\bm{x}\in\mathbb{R}^n$ and $\bm{u}\in\mathbb{R}^m$ denote the system state and input vectors, respectively; $\ell_T$ is the terminal cost function and $\ell$ is the stage cost function; $N$ is the prediction horizon; $\bm{f}(\cdot)$ represents the system dynamics; $\bm{g}(\cdot)$ includes the inequality and equality constraint on $\bm{x}$ and $\bm{u}$; $\bm{g}_T(\cdot)$ is the constraint on the terminal state; $\bm{x}_{\text{op}}$ denotes the state at current operating point.
The transcribed MPC formulation usually results in a nonlinear programming (NLP) problem, due to the nonlinearity from system dynamics and constraints, which can be solved by off-the-shelf solvers~\cite{wachter2006implementation}.

\subsection{Whole Body Control}

Whole body control (WBC) is a dynamic motion controller that tracks the desired task-space acceleration and ground reaction force (GRF) by solving an OCP at current operating point\cite{kim2019highly,khazoom2022humanoid}. 
Within WBC, joint accelerations, joint torques, and GRF are selected as decision variables, while respecting the full-order model dynamics.
With appropriately designed tasks, WBC can be formulated as minimizing a quadratic cost over task acceleration tracking error. 
The general formulation of WBC is given by:
\begin{equation}
    \begin{aligned}
        \min_{\bm{\tau}, \bm{\lambda}, \ddot{\bm{q}}} \quad 
        &\sum_j \|\bm{J}_j \ddot{\bm{q}} + \dot{\bm{J}}_j\dot{\bm{q}} - \bm{a}_j^{\text{des}} \|_{\bm{Q}}^2 \\
        &\quad + \|\bm{\lambda} - \bm{\lambda}^{\text{des}}\|_{\bm{R}_\lambda}^2 + \|\bm{\tau}\|_{\bm{R}_\tau}^2  \\
        \text{s.t. } \quad 
        &\bm{H}\ddot{\bm{q}}
        + \bm{C}
        = \bm{S}_a^\top \bm{\tau}
        + \bm{J}_c^\top \bm{\lambda} \\
        &[\bm{\lambda},\bm{\tau}] \in \mathcal{U}, \\
    \end{aligned}
\end{equation}
where 
$\bm{Q}$, $\bm{R}_{\lambda}$, $\bm{R}_{\tau}$ are the weight matrices for the configuration accelerations, GRF, and torques; $\bm{J}_j$ and $\bm{a}_j^{\text{des}}$ denotes the Jacobian and desired acceleration of the $j$-th task; 
$\bm{q}$ denotes the generalized coordinate. $\bm{H}$ is the mass matrix; 
$\bm{C}$ contains the centripetal, Coriolis, and gravitational terms; 
$\bm{\lambda}$ is the reaction force at the contact points; 
and $\bm{J}_c$ is the corresponding contact Jacobian;
$\bm{S}_a$ is the selection matrix to map actuated joint torques $\bm{\tau}$ to generalized coordinate; 
and $\mathcal{U}$ represent the feasible set for the joint torques and GRF.
The above WBC formulation can be transcribed to a quadratic programming (QP) problem, which can be efficiently solved via QP solvers~\cite{bambade2025proxqp}.

\section{Three Rigid Bodies Model}

This section presents the derivation of the Three Rigid Bodies (TRB) model, a reduced-order model for wheeled bipedal robots that retains the hip roll DoF and supports multiple contact modes.
The modeling procedure, wheel contact kinematics, and the equations of motion (EoM) are detailed in the following subsections.

\subsection{Minimal Coordinate Modeling}
\begin{figure}[tb]
    \centering
    \includegraphics[width=0.95 \linewidth]{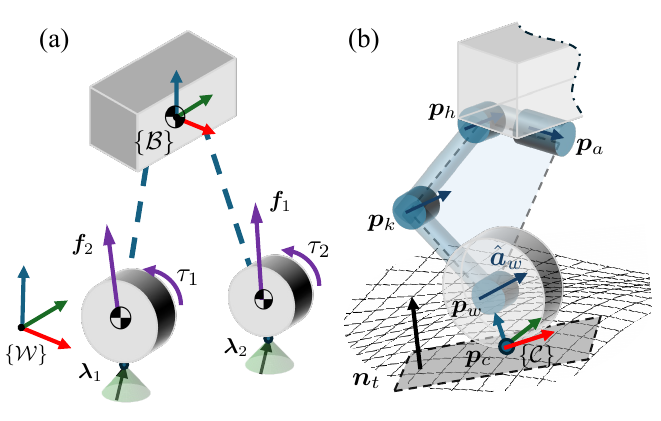}
    \caption{The TRB model. (a) Definition of the world frame \(\{\mathcal{W}\}\), torso body frame \(\{\mathcal{B}\}\), and inputs in the TRB model; 
    (b) Illustration of the leg plane (blue shaded), the associated joint axis positions, and the definition of the contact frame \(\{\mathcal{C}\}\) and the contact point.
    }
    \label{fig:TRBmodel}
\end{figure}

The TRB model abstracts the wheeled bipedal robot as three rigid bodies representing the torso and two wheels, where each wheel rigid body is connected to the torso by a massless, extendable rod that can transmit force and torque.
This simplification relies on the key assumption that the leg mass and inertia are negligible compared to those of the torso and wheels, which is valid for most existing wheeled bipedal robot platforms. Specifically, the leg mass of \textit{Tron1} is less than $10 \%$ of the robot's total mass, which supports our assumption.
Among the three rigid bodies, the torso is modeled as a 6-DoF floating-based rigid body, while the wheels are not completely free due to the constraints imposed by the leg kinematics.
Namely, the rotation axis of each wheel is constrained to remain perpendicular to the corresponding leg plane, as shown in Fig. \ref{fig:TRBmodel}(b). 
The wheel rotation axis is determined by the torso pose and wheel position: 
\begin{equation}
\label{eq:kinematics_constraint}
\begin{aligned}
\bm{a}_w = (\bm{p}_{w} - \bm{p}_{a}) \times (\bm{p}_{a} - \bm{p}_{h}),
\end{aligned}
\end{equation}
where $\bm{a}_w \in \mathbb{R}^3$ denotes a vector along the wheel rotation axis. $\bm{p}_{a}, \bm{p}_{h}, \bm{p}_{w} \in \mathbb{R}^3$ are the position vectors of the hip roll joint, hip pitch joint, and wheel joint, respectively.  All quantities are expressed in the world frame $\{\mathcal{W}\}$ defined in Fig.~\ref{fig:TRBmodel}(a). 

If each wheel were modeled as a 6-DoF floating rigid body, kinematics constraints must be imposed for consistency. 
Such a constrained system is unfavorable for dynamics integration, as classical integration schemes, such as the forward Euler method, can not guarantee constraint satisfaction.
However, the minimum coordinate modeling can describe the pose of each wheel by four variables without additional constraints:
$\bm{q}_{w,i} = [\bm{p}_{w,i}^\top, \phi_{w,i}]^\top \in \mathbb{R}^4$,
where $i \in \{1,2\}$ represent the left and right side, respectively. $\bm{p}_{w,i} \in  \mathbb{R}^3, \  \phi_{w,i} \in  \mathbb{R}$ denote the position vector of wheel's center of mass and the accumulated wheel rotation angle.
The configuration for the TRB model is:
\begin{equation}
    \bm{q} = [\bm{p}_{t}^\top, \bm{\Theta}_{t}^\top, \bm{q}_{w,1}^\top, \bm{q}_{w,2}^\top]^\top,
\end{equation}
where $\bm{p}_{t} \in \mathbb{R}^3$ represents torso position and $\bm{\Theta}_{t} \in \mathbb{R}^3$ for torso orientation, expressed as Euler angles. The control input for the TRB model includes the internal interaction wrenches between the torso and wheels, and the GRF. The interaction wrenches $\bm{u}_{w,i} \in \mathbb{R}^4$ can be described as:
\begin{equation}
    \bm{u}_{w,i} = [\bm{f}_{i}^\top, \tau_{i}]^\top,
\end{equation}
where $\bm{f} \in \mathbb{R}^3$ is the interaction force and $\tau \in \mathbb{R}$ denotes the rotational torque acting along with the wheel rotation axis.
The input vector of the system is defined as:
\begin{equation}
\bm{u} = [\bm{u}_{w,1}^\top, \bm{u}_{w,2}^\top, \bm{\lambda}_1^\top, \bm{\lambda}_2^\top]^\top ,
\end{equation}
where $\bm{u} \in \mathbb{R}^{14}$ is the system input vector and $\bm{\lambda} \in \mathbb{R}^3$ denotes the GRF to the wheel.

\subsection{Wheel Contact Kinematics}
\label{sec:wheel_contact_kinematics}
A contact frame \{$\mathcal{C}$\}, as shown in Fig~\ref{fig:TRBmodel}(b), is defined to describe the wheel contact kinematics. The $y$-axis of this frame is parallel to the rotation axis of the wheel $\bm{a}_w$; the $z$-axis points from the contact point toward the wheel’s center of mass. Given the wheel joint axis and the terrain normal vector $\hat{\bm{n}}_t \in \mathbb{R}^3$, the rotation matrix of the contact frame $R_{sc} = [\hat{\bm{x}}, \hat{\bm{y}}, \hat{\bm{z}}]$ can be determined as:
\begin{equation}
\label{eq:contact_frame}
\hat{\bm{x}} = \frac{\hat{\bm{a}}_{w} \times \hat{\bm{n}}_{t}}
    {\left\| \hat{\bm{a}}_{w} \times \hat{\bm{n}}_{t} \right\|}, \quad
\hat{\bm{y}} = \hat{\bm{a}}_w, \quad
\hat{\bm{z}} = \hat{\bm{x}}\times\hat{\bm{y}},
\end{equation}
where $\hat{\cdot}$ denotes the normalized vector. The contact point position $\bm{p}_c \in \mathbb{R}^3 $  can be written as $ \bm{p}_c =  \bm{p}_w - r_w\hat{\bm{z}}$, while
the contact velocity $\bm{v}_c \in \mathbb{R}^3 $ satisfies $\bm{v}_c = \bm{v}_w - \bm{\omega}_w \times r_w \hat{\bm{z}}$. The angular velocity of the wheel $\bm{\omega}_w \in \mathbb{R}^3$ can be computed via forward kinematics with contributors from torso and joint velocity, whose value can be approximated as $\dot{\phi}_w$, assuming the out-of-axis rotation is negligible. The contact velocity is:
\begin{equation}
\label{eq:contact_vel}
    \bm{v}_c \approx \dot{\bm{p}}_w - \dot{\phi}_w r_w\hat{\bm{x}} .
\end{equation}
The out-of-axis rotation consists of the hip roll joint velocity and the torso angular velocity. Note that the torso yaw rate has little effect on the contact velocity, since it is largely aligned with $\hat{\bm{z}}$. Therefore, this assumption is valid in most cases where the hip roll, the torso pitch, and torso roll velocities remain small.
With this explicit expression of the contact velocity, the TRB model supports multiple contact modes, including no-contact, rolling contact, and slipping contact, by imposing corresponding constraints on the wheel contact velocity.

\subsection{TRB Model Equations of Motion}
The dynamical equations for the torso are:
    \label{eq:torso_dyn}
    \begin{align}
    \ddot{\bm{p}}_t &= \frac{1}{m_t}\sum_{i=1}^2 \bm{f}_i - \bm{g}, \label{eq:torso_dyn_translation} \\ 
     \dot{\bm{\omega}}_t  &= \bm{I_t}^{-1}\left(-\bm{\omega}_t\times\bm{I}_t \bm{\omega}_t + \sum_{i=1}^2 ( \bm{r}_i \times \bm{f}_i + \tau_i\hat{\bm{y}}_i)\right)\label{eq:torso_dyn_orien},
    \end{align}
where $m_t \in \mathbb{R}$, $\bm{\omega}_t \in \mathbb{R}^3$ and $\bm{I}_t \in \mathbb{R}^{3\times3}$ denote the scalar mass, angular velocity, and inertia matrix of the torso, expressed in $\{\mathcal{W}\}$. 
The vector $\bm{r}_i =  \bm{p}_{w,i} - \bm{p}_t $ represents the position vector from the torso to the $i$-th wheel. $\bm{g} \in \mathbb{R}^3$ is the acceleration of gravity.
We use the Euler angles for the orientation representation, whose time derivative could be expressed as 
\begin{equation}
    \dot{\bm{\Theta}}_t=\mathbf{T}(\bm{\Theta})\bm{\omega}_t,
    \label{eq:euler_rate}
\end{equation}
where $\mathbf{T}(\cdot)$ denotes the transformation matrix that converts angular velocity to the time derivative of Euler angles\cite{di2018dynamic}. Combining the \eqref{eq:torso_dyn_orien} and \eqref{eq:euler_rate}, $\ddot{\bm{\Theta}}_t$ could also be expressed as a function of the TRB state and input.
The EoM for the wheel rigid body is:
\begin{equation}
     \label{eq:wheel_dyn}
        \ddot{\bm{q}}_{w,i}  = \bm{H}_w^{-1}( \bm{u}_{w,i} + \bm{S}_{w,i}^\top\bm{\lambda}_{i} -\bm{C}),
\end{equation}
where $\bm{S}_{w,i}= [\,\bm{I}_{3\times3},\ -r_w\hat{\bm{x}}_i\,] \in \mathbb{R}^{3\times4}$ denotes the wheel contact Jacobian matrix. $\bm{C} = [\,\bm{g};\,0\,] \in \mathbb{R}^4$ accounts for gravity, and the wheel mass matrix is defined as $\bm{H}_w = \operatorname{diag}(m_w, m_w, m_w, I_w) \in \mathbb{R}^{4\times4}$, 
where $m_w \in \mathbb{R}$ is the mass of the wheel and $I_w \in \mathbb{R}$ is the moment of inertia about the wheel’s joint axis.  By concatenating the TRB model configuration $\bm{q}$ and its time derivative $\dot{\bm{q}}$ into the system state $\bm{x} = [\bm{q}^\top,\dot{\bm{q}}^\top]^\top$, the EoM of TRB dynamics can be written in the form of $\ddot{\bm{q}} = \bm{f}(\bm{x},\bm{u})$ by combining above equations.

\begin{figure*}[tb]
    \centering
    \includegraphics[width=1\linewidth]{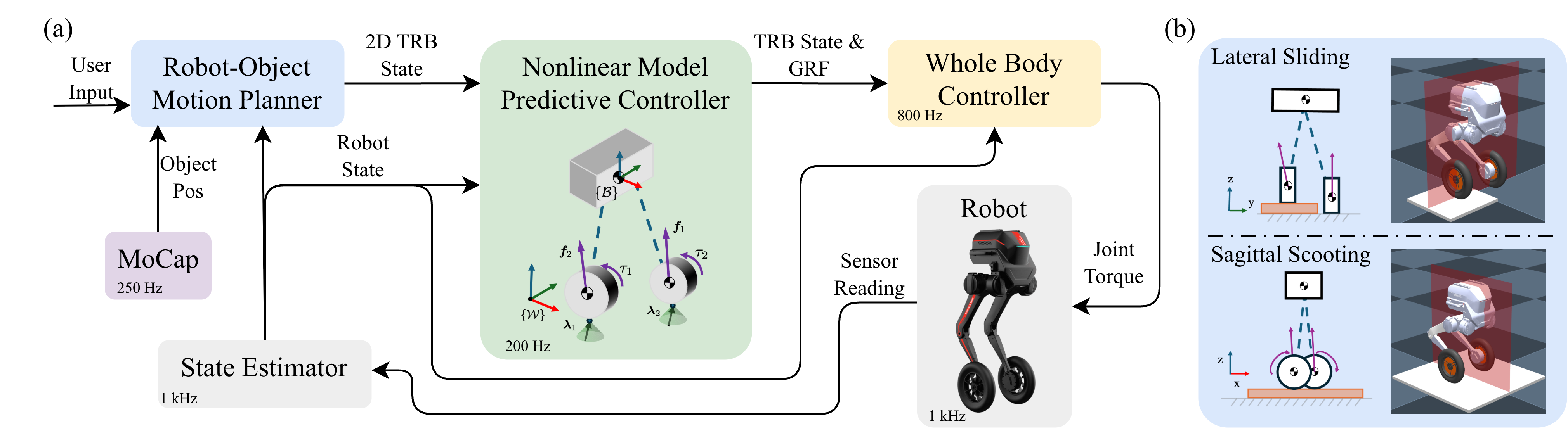}
    \caption{Overview of the proposed hierarchical control framework. (a) User inputs are sent to the robot-object motion planner, which generates a 2D TRB state reference. The reference and the estimated robot state are provided to the NMPC.
    The NMPC computes the desired TRB states and interaction GRF, which are passed to the whole-body controller. The state estimator based on sensor data from robot hardware is streaming at 1 kHz. The MoCap system detects the object pose at 250 Hz, which is only used in pedipulation experiment. (b) The 2D TRBO models are used in the robot-object motion planner.}
    \label{fig:hierarchical-controller-diagram}
\end{figure*}

\section{Hierarchical Control Framework}

The proposed hierarchical control framework aims to enable the wheeled bipedal robot to perform dynamic mobile pedipulation tasks that involves object sliding. 
Within the framework, a robot-object motion planner generates the reference trajectory to initiate the stick-slide transition, which is tracked by the NMPC. A whole-body controller is employed to enhance the robustness of the controller.

\subsection{Robot-Object Motion Planner} \label{sec:robot_object_motion_planner}
To simultaneously capture the underactuated robot dynamics and object dynamics for robot-object motion planning, the TRB model is combined with an object model, forming the TRB-Object (TRBO) model. The object is subjected to forces from the wheel and ground contact:
\begin{equation}
    \ddot{\bm{p}}_o = \frac{1}{m_o}(\bm{f}_o - \bm{\lambda}_{r,xy}), 
\end{equation}
where $\bm{p}_o = [x_o, y_o]^{\top}$ denotes object position; $\bm{f}_o$ is the frictional force from the object-ground contact and $\bm{\lambda}_{r}$ represent the object-wheel contact force applied on the wheel. The object is assumed to have no acceleration in the $z$ direction and remains in contact with the ground, hence:
\begin{equation}
    N = \lambda_{r,z} + m_o g,\label{eq:object_Fn}
\end{equation}
where $N$ denotes the normal supporting force from the ground.
Object sliding is achieved by imposing a phase-dependent object-ground contact constraint:
\begin{equation}
\begin{cases}
\left\| \bm{v}_{o,k} \right\| = 0, \  \|\bm{f}_{o,k}\| \leq \mu_{o} N_k,  & s_k = 0\\[3pt]
\left\| \bm{v}_{o,k} \right\| > 0, \ \bm{f}_{o,k} = -\mu_o N_k\hat{\bm{v}}_{o,k}, & s_k = 1,
\end{cases}
\label{eq:object_coulomb_law}
\end{equation}
where the $\bm{v}_{o}$ is the planar velocity of the object; $\mu_o$ is the friction coefficient of ground-object contact;
$s_k \in \{0,1\}$ is a pre-defined Boolean variable indicating whether sliding occurs at the $k$-th time step, which is fixed and can not be changed by the optimizer. 

Within the motion planner, the TRBO model and contact constraints are simplified into task-dependent planar models to reduce solve time. The lateral sliding mode adopts the frontal plane 2D TRBO model, while the scooting mode adopts the sagittal plane model, as shown in Fig.~\ref{fig:hierarchical-controller-diagram}(b).
A trajectory optimization problem is formulated based on the 2D TRBO model using direct collocation\cite{hargraves1987direct}. 
Input limit constraints are also incorporated into the optimization problem. 
The resulting OCP is transcribed into an NLP problem and solved using off-the-shelf solvers. The output of this motion planner consists of the desired robot-object state trajectory and GRF profiles, which serve as references for the NMPC module.

\subsection{Nonlinear Model Predictive Control}
Given the reference generated by the robot-object motion planner, an NMPC module based on the TRB model is developed to coordinate wheel–leg motion and precisely regulate contact forces. This subsection presents the NMPC formulation, including the cost function design, TRB dynamics constraints, wheel contact constraints, and TRBO extension.

\subsubsection{Cost Function}
To minimize the tracking error and control effort, the cost function is defined as:
\begin{equation}
\ell_(\bm{x}, \bm{u}) =
\left\| \bm{x} - \bm{x}^{\mathrm{des}} \right\|_{Q}^{2} + \left\| \bm{u} - \bm{u}^{\mathrm{des}} \right\|_{R}^{2}.
\end{equation}
The superscript $(\cdot)^{\mathrm{des}}$ indicates desired values, which come from the robot-object motion planner.
For the terminal cost in the MPC controller, we approximate the system’s value function as
\begin{equation}
    \ell^{T}(\bm{x}) 
    = \left\| \bm{x} - \bm{x}^{\mathrm{des}} \right\|_{\bm{S}}^{2},
\end{equation}
where $\bm{S}$ is the solution to the associated algebraic Riccati equation, computed offline using the linearized dynamics around the nominal state and input. 
This weighted quadratic cost approximates the infinite-horizon cost-to-go under a simplified linear model with relaxed constraints. 
The terminal cost effectively increases the controller's prediction horizon and robustness without incurring significant computational overhead \cite{li2021model}.

\subsubsection{TRB Dynamics Constraint}
\label{sec:trb_dynamics}

The dynamics constraints are formulated by discretizing the system dynamics using a semi-implicit Euler integration scheme with a time step $\delta t$:
\begin{equation}
    \begin{aligned}
        \dot{\bm{q}}_{k+1} &= \dot{\bm{q}}_k + \bm{f}(\bm{x}_k,\bm{u}_k) \delta t \\
        \bm{q}_{k+1} &= \bm{q}_k + \dot{\bm{q}}_{k+1}\delta t .
    \end{aligned}
    \label{eq:TRB_dyn_constraint}
\end{equation}

\subsubsection{Wheel Contact Constraint}

For the wheel-ground contact, the wheel contact velocity is zero under non-slip assumption.
According to the Eq. \eqref{eq:contact_vel}, the time derivative of contact velocity is:
\begin{equation}
    \dot{\bm{v}}_c = \ddot{\bm{p}}_w - r_w \ddot{\phi}_w \hat{\bm{x}} - r_w \dot{\phi}_w \dot{\hat{\bm{x}}} .
\end{equation}

When the wheel maintains rolling contact with the object's top surface without slip, the contact constraint is written as:
\begin{equation}
\label{eq:contact_constraint_wheel}
\left\{
\begin{aligned}
 \dot{\bm{v}}_{c_i} &= 0 
        && \text{if } c_i \in \mathcal{C}_g \\
    \dot{\bm{v}}_{c_i} & = \bm{a}_o
        && \text{if } c_i \in \mathcal{C}_o,
\end{aligned}
\right. 
\end{equation}
where $c_i$ denotes the contact point of the $i$-th leg and $\bm{a}_0 \in \mathbb{R}^3$ represents the object acceleration.
The sets $\mathcal{C}_g$ and $\mathcal{C}_o$ indicate ground–contact points and object–contact points, respectively. 
Meanwhile, the reaction forces exerted on the robot satisfy the friction–cone constraint:
\begin{equation}
    \begin{aligned}
    \bm{\lambda}_i &\in  \{\bm{\lambda} \  | \ 0\leq \lambda_{z}\leq \lambda_{\text{max}},\, \|\bm{\lambda}_{xy}\| \leq \mu_i \lambda_{z} \},
    \end{aligned}
    \label{eq:friction_cone_constraint}
\end{equation}
where $\mu_i \in \{\mu_{w,g},\mu_{w,o}\}$ is the friction coefficient between the wheel on the $i$-th leg and the contacting surface. $\mu_{w,g}$ and $\mu_{w,o}$ are the friction coefficients of wheel-ground contact and wheel-object contact.
$\lambda_{\text{max}}$ is the maximum vertical GRF.

\subsubsection{Swing Phase Constraint}
During the swing phase of a stepping motion, the wheel contact velocity is unconstrained, and the corresponding GRF vanishes:
\begin{equation}
    p_{c,z} > 0,\quad \bm{\lambda} = 0.
\end{equation}

\subsubsection{TRBO NMPC Extension}\label{sec:TRBO_extension}
For tasks requiring accurate object-state tracking, the TRB-based NMPC can be extended to a TRBO-based NMPC, by incorporating the object dynamics into the model. Here, the object model follows the same point-mass assumption used in the robot-object motion planner described in Section~\ref{sec:robot_object_motion_planner}. 

To enable online execution of the TRBO-based NMPC, the object model is simplified to one-dimensional along the desired sliding direction. Taking the object sliding task along the $y$-axis as an example, the object position $y_o \in \mathbb{R}$ and friction force $f_{o,y} \in \mathbb{R}$ along the $y$-axis are selected as the object configuration and input. The dynamics are:
\begin{equation}
    \ddot{y}_o = \frac{1}{m_o}(f_{o,y} - \lambda_{r,y})
    \label{eq:ob_1d_dyn}
\end{equation}
Combining the object state with TRB state, the TRBO system state and input are defined as
\begin{equation}
    \bm{x}_u = [\bm{x}^\top,y_o,\dot{y}_o]^\top \quad \bm{u}_u=[\bm{u}^\top,f_{o,y}]^\top.
\end{equation}

Then, a sliding constraint is added to the NMPC to enforce Coulomb friction complementarity constraint. Given the desired sliding direction $\hat{d}\in\{1,-1\}$ along the $y$-axis, the sliding constraint is compactly represented as:
\begin{equation}
\begin{cases}
\dot{y}_{o,k} = 0, \quad  {f}_{o,y,k} \leq \mu_{o} N_k,  & s_k = 0\\[3pt]
\hat{d}\dot{y}_{o,k} > 0, \quad {f}_{o,y,k} = - \hat{d}\mu_o N_k, & s_k = 1,
\end{cases}
\label{eq:object_coulomb_law_1D}
\end{equation}
This simplified constraint substantially reduces the complexity of the optimization problem, which is critical because the object dynamics are directly incorporated into the NMPC and must be solved rapidly online.

\section{Result}

\subsection{System Setup}

This work uses the \textit{Tron1} wheeled bipedal robot, which weighs approximately 20 kg with 8 actuated leg joints.
MuJoCo\cite{todorov_mujoco_2012} is used as the simulation environment.
The optimal controller formulation is implemented in CasADi~\cite{andersson2019casadi} and executed on a desktop computer equipped with a 13th~Gen Intel Core i7\textendash13700 CPU, communicating with the robot middleware via LCM~\cite{huang2010lcm}.
With this setup, the TRB MPC controller runs at 175$\pm$11 Hz, enabled by the boosted solving time of \textit{Fatrop} solver~\cite{vanroye2023fatrop}. 
WBC tasks are designed to track torso linear and angular accelerations, wheel center-of-mass (CoM) positions and velocities, and dampen contact velocities. The full-order dynamics are computed via \textit{Pinocchio}~\cite{carpentier2019pinocchio} from URDF. WBC is formulated via CasADi as well, and the solved joint torques are commanded to the robot at 800 Hz.
The state estimator module runs a Kalman Filter, which fuses onboard IMU and joint encoders' readings to estimate the pose of the robot torso at 1 kHz.
The Vicon motion capture (MoCap) system locates the position of the object for the motion planner to correct the plan with object state feedback, which is only used in pedipulation experiment.

\subsection{Locomotion Experiment}
\begin{figure}[htb]
    \centering
    \includegraphics[width=0.95\linewidth]{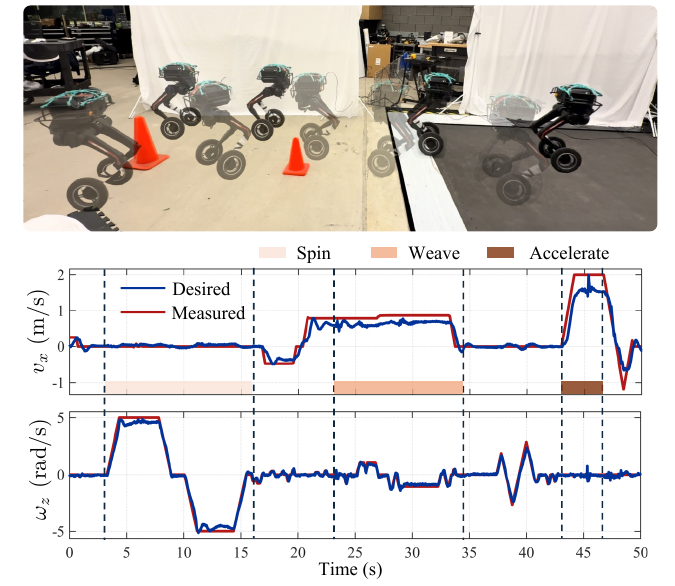}
    \caption{Velocity tracking experiment. Top: snapshots during the combined locomotion task, including spinning, weaving, and forward acceleration. Bottom: the desired and measured velocities from proprioceptive state estimator, where the shaded regions indicate the task execution intervals.}
    \label{fig:vel_tracking}
\end{figure}
To verify the effectiveness of the proposed framework, we deployed the controller on the hardware platform and commanded the robot to track velocity commands from a human-operated joystick. We designed a combined locomotion experiment to evaluate velocity tracking in multiple directions without the motion capture system. In this experiment, the robot first performed rapid in-place spinning, then followed a weaving trajectory around two obstacle cones, and finally executed fast forward acceleration. As shown in Fig.~\ref{fig:vel_tracking}, the robot successfully completed all tasks and demonstrated reliable velocity tracking performance.

\subsection{Pedipulation Experiments}
We conducted two pedipulation experiment, lateral sliding and scooting. In the lateral sliding experiment, the robot either pushes away or retrieves an object.
In the scooting experiment, as shown in Fig.~\ref{fig:scooting}, the robot leverages its own dynamics to initiate intermittent object sliding.

\subsubsection{Lateral Sliding}
\begin{figure}[tb]
    \centering
    \includegraphics[width= 0.95\linewidth]{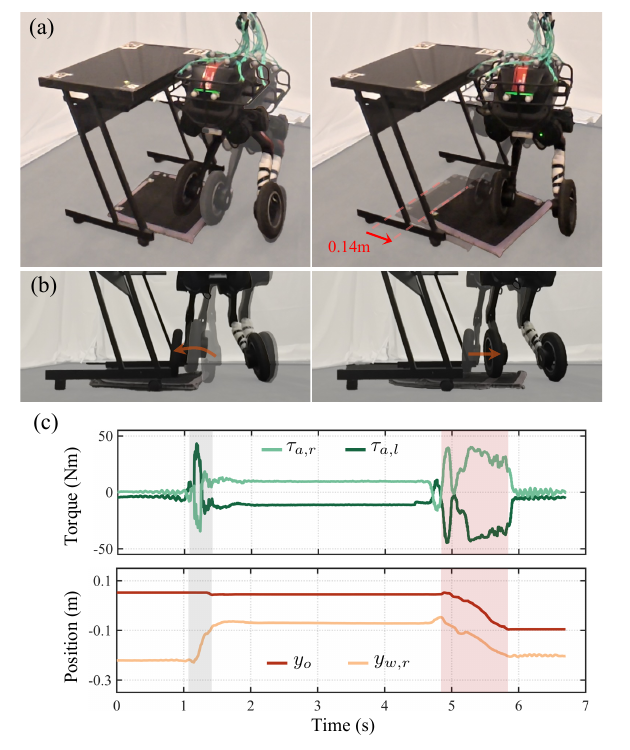}
    \caption{Wheeled bipedal robot retrieves a planar object from under a desk using its wheeled legs. (a) Isometric view of the stepping and sliding, where the red arrows indicate the object displacement. (b) Front view, where the orange arrows indicate the direction of the wheel motion. (c) The hip roll joint torque and position trajectory of the object and the pedipulating wheel. $\tau_{a,l}$ and $\tau_{a,r}$ are the left and right hip roll joint motor torque commands. $y_o$ and $y_{w,r}$ are the lateral position of object and pedipulating wheel. The stepping and sliding intervals are highlighted in gray and red, respectively.}
    \label{fig:retrieving}
\end{figure}
During the lateral sliding experiment, shown in Fig.~\ref{fig:retrieving}, the wheeled bipedal robot steps sideways onto the object and retrieves it from under a desk.
The hip roll joint actuator plays an important role in this task. As shown in the torque profile in Fig.~\ref{fig:retrieving}(c), the right hip roll joint motor swings the stepping wheel laterally while the other compensates for gravity and maintain balance. During the retrieval motion, both actuators  generates torque to produce a contact force in the lateral direction. 

\subsubsection{Scooting}
During the scooting experiment, both wheels of the robot are on a large planar object.
As presented in Fig.~\ref{fig:scooting}(a), the robot periodically move forward and brake, inducing the object to intermittently slide forward.
The GRF trajectories generated by the NMPC, shown in Fig.~\ref{fig:scooting}(b), provide insight into the scooting motion. The robot first accelerates and then rapidly brakes to generate a large forward contact force that initiates object sliding. During the braking phase, the vertical force is reduced to decrease the static friction at the object--ground contact interface.
This cycle can be executed repeatedly to achieve large object displacements.
In the scooting experiment shown in Fig.~\ref{fig:scooting}, the robot performs eight scooting cycles, resulting in a total displacement of 0.228 m.

\subsubsection{Object Parameter Sweep in Simulation}
\begin{figure}[htb]
    \centering
    \includegraphics[width=0.75\linewidth]{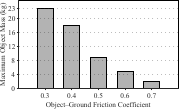}
    \caption{Maximum object mass that Fig.~\ref{fig:scooting} can slide in the scooting mode under varying object-ground friction coefficient. Simulation results.}
    \label{fig:scooting_object_para}
\end{figure}

The object parameters used in the hardware experiment are summarized in Table~\ref{tab:object_params}. To further evaluate the robustness of the controller with respect to the object properties, an object parameter sweep is performed on the scooting controller in simulation.
Fig.~\ref{fig:scooting_object_para} shows the maximum object mass that \textit{Tron1} could scoot under varying object-ground friction coefficients. The scooting motion is considered feasible if the displacement over one scooting cycle exceeds $1$ cm. The wheel-ground and wheel-object friction coefficients are fixed as $0.9$. 
The result shows that \textit{Tron1} can initiate sliding motion for objects with mass up to 23 kg when the friction coefficient is 0.3, which is heavier than the robot’s own mass, demonstrating the potential of this motion skill for manipulating large and heavy objects in industrial applications.

\begin{table}[htb]
    \caption{Hardware Experiments Parameters}
    \label{tab:object_params}
    \centering
    \begin{tabular}{l c c}
        \toprule
        \textbf{Parameters} & \textbf{Scooting} & \textbf{Lateral Sliding} \\
        \midrule
        Object Mass & 4 kg & 1 kg \\
        Wheel/Obj Friction Coeff & 1.0 & 1.0 \\
        Obj/Ground Friction Coeff & 0.3 & 0.3 \\
        Wheel/Ground Friction Coeff & 0.8 & 0.8 \\
        \bottomrule
    \end{tabular}
\end{table}

\subsubsection{TRBO-based NMPC Experiment in Simulation}
To quantify the object state tracking performance in our framework, we conduct simulation tests on the lateral sliding task using both TRB-based and TRBO-based NMPC. In this task, we omit the stepping process and initialize the robot in contact with the object. Then the controller drives the robot to push the object farther away from the torso. Moreover, to assess the robustness to modeling uncertainty, we conducted several experiments on different actual object masses while keeping the estimated object mass in TRBO model fixed.

\begin{figure}[htb]
    \centering
    \includegraphics[width=0.95\linewidth]{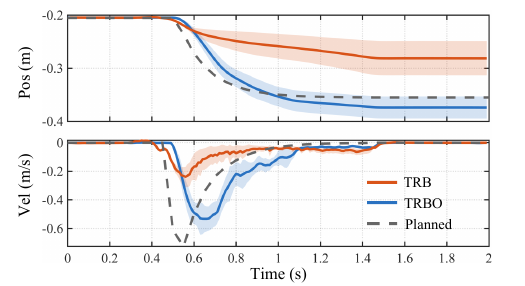}
    \caption{Object position and velocity along the sliding direction in the simulated lateral sliding task with inaccurate object-mass estimation. The solid line and shaded region represent the mean value and standard deviation across the successful trials.}
    \label{fig:trbo_simu}
\end{figure}

The object trajectories are shown in Fig.~\ref{fig:trbo_simu}. In this experiment, the planner and NMPC use a fixed nominal object mass of $2~\mathrm{kg}$, while the actual mass varies from $1~\mathrm{kg}$ to $3~\mathrm{kg}$ across 6 trials. Despite this mass mismatch, the TRBO-based NMPC succeeds in all trials with lower tracking error, achieving a $100\%$ success rate $(6/6)$. In contrast, the TRB-based NMPC succeeds in only $66\%$ of the trials $(4/6)$, highlighting the limitation of excluding object dynamics from the predictive model.

\section{Conclusion And Future Work}
This letter presents a hierarchical control framework that enables a wheeled bipedal robot to perform object-sliding task via mobile pedipulation. 
The proposed TRB dynamical model enables the robot to interact with objects in both longitudinal and lateral directions due to its inclusion of the hip roll DoF.
The robot-object motion planner captures the unified robot-object dynamics, including object stick-slip transition and the robot's underactuated dynamics, by formulating a trajectory optimization problem based on a planar TRBO model.
Then, the NMPC controller robustly tracks the desired trajectory generated by the robot-object motion planner through coordinated wheel-leg motion replanning and precise contact-force regulation.
A WBC further refines the input and state command from NMPC using a full-order model. 
Locomotion experiments are conducted in hardware and demonstrate stable control performance. Among pedipulation experiments, the robot successfully retrieves a $1~kg$ object from under a desk via stepping followed by lateral sliding.
Scooting motion is demonstrated by sliding a $4~kg$ object over a distance of $0.228~m$. 
Simulation results further indicate the controller's capability for heavy object transportation, achieving sliding pedipulation on an object with a mass of up to $23$ kg. The TRBO-based NMPC extension also achieves low object-state tracking error and demonstrates robustness to model mismatch.
Overall, the results demonstrate robust locomotion control and effective handling of dynamic robot-object interaction. 

Regarding the limitations of this work, first, the pedipulation experiments currently rely on motion capture for object-state estimation, which limits deployment outside instrumented laboratory environments. Second, the object is modeled as a point mass, which does not capture object rotation, distributed contact geometry, or contact-patch effects. Third, the current controller assumes a pre-defined contact sequence for the stepping and sliding phases. Future work will address onboard object-state estimation, richer object and contact models, and online contact-sequence adaptation toward more autonomous pedipulation.


\bibliographystyle{IEEEtran}
\bibliography{reference}

\vfill

\end{document}